\documentclass{article}

\usepackage{microtype}
\usepackage{graphicx}
\usepackage{subfigure}
\usepackage{booktabs} 

\usepackage{color}

\usepackage{hyperref}

\usepackage{amssymb}
\usepackage{amsmath}
\usepackage{bm}

\DeclareMathOperator*{\minimize}{min}

\usepackage[accepted]{icml2020}

\icmltitlerunning{Meta-learnt priors slow down catastrophic forgetting in neural networks}

\begin{document}

\twocolumn[
\icmltitle{Meta-learnt priors slow down catastrophic forgetting in neural networks}



\icmlsetsymbol{equal}{*}

\begin{icmlauthorlist}
\icmlauthor{Giacomo Spigler}{equal,uvt}
\end{icmlauthorlist}

\icmlaffiliation{uvt}{Department of CSAI, Tilburg University, Tilburg, Netherlands}

\icmlcorrespondingauthor{Giacomo Spigler}{g.spigler@tilburguniversity.edu}

\icmlkeywords{sequential learning, sequential task learning, continual learning, meta-learning, catastrophic forgetting, Baldwin effect, neuro-evolution}

\vskip 0.3in
]



\printAffiliationsAndNotice{\icmlEqualContribution} 

\begin{abstract}
Current training regimes for deep learning usually involve exposure to a single task / dataset at a time. Here we start from the observation that in this context the trained model is not given any knowledge of anything outside its (single-task) training distribution, and has thus no way to learn parameters (i.e., feature detectors or policies) that could be helpful to solve other tasks, and \textit{to limit future interference with the acquired knowledge}, and thus catastrophic forgetting.

Here we show that catastrophic forgetting can be mitigated in a meta-learning context, by exposing a neural network to multiple tasks \textit{in a sequential manner} during training. Finally, we present SeqFOMAML, a meta-learning algorithm that implements these principles, and we evaluate it on sequential learning problems composed by Omniglot and MiniImageNet classification tasks. 
\end{abstract}

\section{Introduction}

Current deep learning systems are very effective at solving specific, narrow tasks, and can already beat humans in many contexts (e.g., playing chess, GO \cite{alphago}, and object recognition \cite{human_level_imagenet}). However, they are still limited in the generality of their application, and learning in artificial agents remains very different from biological learning. Two problems in particular need to be addressed: the algorithms should be more sample efficient, so that they can learn from few data points instead of requiring enormous datasets, and the models used should be able to learn continually and incrementally, without forgetting their previously learnt skills and instead possibly using them to speed-up future learning.

Here we suggest that \textit{one of the limitations that exacerbates these two problems is that current deep learning training regimes involve exposure to only a single task}. The problem stems from the fact that a model trained on a single task cannot extract any knowledge of anything that is not in its training dataset. For example, an agent trained to play the Pong Atari game, which consists of simple shapes, and has no colored objects, has no need to learn feature detectors for color or complex shapes, even though from our perspective it would seem useful to have them, in light of generality and to better adapt to other tasks that require complex vision.

This limitation becomes particularly important in the context of \textit{sequential task learning}, whereby a model, for example a neural network, is required to learn to perform well on tasks that are presented sequentially, one at a time, while retaining knowledge of the previously observed ones. In this context, if the model is trained on a single task, it has no way to determine what parameters (yielding features or behaviors) could be useful for future tasks, resulting in severe overfitting from the perspective of future tasks the model could observe, and thus not only limiting the potential transfer between tasks, but also making future learning harder. Further, because the model is only trained on a single task, it is given no knowledge about the fact that it will be later required to learn other tasks, and that it will need to retain the previously gained knowledge.

\textit{In this paper,} we focus on the problem of catastrophic forgetting, that is the tendency of biological and artificial neural networks to rapidly forget previously learnt knowledge upon learning new information \cite{mccloskey1989catastrophic, goodfellow2013empirical}. \textit{We propose to address the problem by extending a meta-learning framework to pre-train models using sequences of tasks, rather than individual tasks, to learn good, innate priors that are optimized for transfer and adaptability to tasks from a given distribution and for resistance to catastrophic forgetting.} We further formalize the problem explicitly in the case of sequential task learning.

Meta-learning \cite{metalearning_schmidhuber1987, metalearning_bengio1992optimization, ml_graddesc_bygraddesc_2016}, that is the problem of building systems that ``learn to learn'', is particularly well suited for the purpose, as it has been already extensively used to optimize models for fast adaptation to new tasks \cite{maml}, addressing the first part of the problem. For example, meta-learning algorithms that optimize the initial model parameters using a distribution of tasks instead of individual tasks can learn good \textit{priors} for novel tasks that are similar to those belonging to the meta-training distribution, enabling fast learning on such tasks.

It is interesting to observe that this type of meta-learning shows an affinity with biological nervous systems. Specifically, biological agents are not born as a ``blank state'', but rather come with innate ``evolutionary priors'' as genome-encoded knowledge that affects the agents' broad initial brain connectivity, leads to the development of specific feature detectors (especially in the early sensory cortices), and to innate behaviors and skills. The interplay between biological evolution and within-lifetime learning and adaptation has been explored in biology \cite{baldwin_effect_original} and in connectionism / neural networks \cite{baldwin_effect_neuralnetworks} as the \textit{Baldwin effect}, and has recently been connected to meta-learning \cite{baldwinian_meta_learning}.

Within the scope of this paper, the connection between meta-learning and biological evolution is developed with a further observation: evolution works with biological agents that experience more than a single task, and in particular \textit{experience tasks throughout an extended time} (their whole life), over which it is advantageous to remember previously learnt skills. Thus the \textit{fitness / objective optimized by evolution} is not the performance over a single task, nor only over a task distribution (as in previous meta-learning methods or Baldwinian meta-learning \cite{baldwinian_meta_learning}), but rather it correlates with the capacity of an agent or model to learn all tasks in a sequence of tasks \emph{quickly} and to \emph{remember them}. This can be made implicit in reinforcement learning settings where an agent needs to solve multiple tasks within the same environment and episode (``lifetime''), and so it is usually implicitly exploited in evolution-based solutions for catastrophic forgetting / continual learning (e.g., \cite{ellefsen2015neural, soltoggio2018born}), or it can be explicitly optimized to learn good priors (in this case, initial model parameters) that help sequential task learning. 

Our main specific contributions are as follows: first, we formalize the problem of continual learning in a meta-learning context based on sequences of tasks, rather than individual tasks. Second, we introduce a variant of FOMAML (First Order MAML) \cite{maml}, \textbf{SeqFOMAML}, that performs meta-learning using meta-batches composed of sequences of tasks rather than individual tasks, and we show that it can be used to reduce the amount of catastrophic forgetting during sequential task learning, and to encourage positive forward and backward transfer between tasks. Finally, we test the performance of the algorithm on sequential learning problems built from Omniglot and MiniImageNet classification tasks, and we show that even traditional meta-learning can achieve a degree of robustness to catastrophic forgetting in multi-headed networks, due to its learning of good reusable features, although its performance is more limited than the proposed SeqFOMAML.  

\section{Theory}
\label{sec:theory}

We first introduce the problem setting using notation from \cite{reptile} and \cite{maml}.

\textbf{Tasks.} Throughout the manuscript we focus on \textit{tasks} as specific supervised-learning problems, but extension to other problems is straightforward. Each task $\mathcal{T}_i$ is associated with a loss function $L_i$.

\textbf{MAML and First-Order MAML (FOMAML).} MAML (Model-Agnostic Meta-Learning) \cite{maml} solves the meta-learning problem by learning a weights initialization that aids transfer to tasks within the meta-training distribution. MAML does not make assumptions on the model aside from it being parameterized by parameters $\boldsymbol{\phi}$, with respect to which gradients can be computed. Meta-learning is then framed as an optimization problem with respect to the initial parameters $\boldsymbol{\phi}$, to minimize the expected loss achieved after training on a task $\tau$ for a number $k$ of updates:
\begin{equation*}
      \minimize_{\boldsymbol{\phi}} \mathbb{E}_\tau \left[ L_{\tau,\text{test}} \left( U_{\tau,\text{train}}^k(\boldsymbol{\phi}) \right) \right]
\end{equation*}
where $U_{\tau,\text{train}}^k$ is a training operator that updates $\boldsymbol{\phi}$ for $k$ times using data sampled from the task-specific \textit{training} set (e.g., usually $U$ is a single step of gradient descent on, $\boldsymbol{\phi}^1 = U_{\tau,\text{train}}^1(\boldsymbol{\phi}) = \boldsymbol{\phi} - \alpha \nabla_{\boldsymbol{\phi}} L_{\tau,\text{train}}(\boldsymbol{\phi})$ \cite{maml,reptile}). The optimization problem can be solved by gradient descent:
\begin{align*}
    g_{\text{MAML}} &= \nabla_{\boldsymbol{\phi}} \mathbb{E}_\tau \left[ L_{\tau,\text{test}}\left(U_{\tau,\text{train}}^k(\boldsymbol{\phi})\right) \right] \\
    &= \mathbb{E}_\tau \left[ U_{\tau,\text{train}}^{'k}(\boldsymbol{\phi}) L_{\tau,\text{test}}'(\boldsymbol{\tilde{\phi}}) \right]
\end{align*}

where $\boldsymbol{\tilde{\phi}} = U_{\tau,\text{train}}^k(\boldsymbol{\phi})$, and $U_{\tau,\text{train}}^{'k}(\boldsymbol{\phi})$ is the Jacobian $\frac{\partial \boldsymbol{\tilde{\phi}}}{\partial \boldsymbol{\phi}}$.
As computing the Jacobian of the update operator is too expensive for all but small $k$, a first-order variant of MAML (FOMAML) has been proposed that approximates it with the Identity matrix, and was found to empirically achieve comparable performance. With a first-order approximation, the meta-gradient becomes:
\begin{equation*}
    g_{\text{FOMAML}} = \mathbb{E}_\tau \left[ L_{\tau,\text{test}}'(\boldsymbol{\tilde{\phi}}) \right]
\end{equation*}


\textbf{Seq(FO)MAML : MAML on sequences of tasks.} We extend MAML \cite{maml} to use task sequences instead of individual tasks, with the objective of both (1) allowing fast learning of tasks in the sequence, and (2) limiting any interference between tasks and thus catastrophic forgetting.

Like MAML, SeqFOMAML acts in two steps. First, the initial parameters of a model are optimized via meta-training as described below. Then the model can be used in practice, by continually updating its parameters using plain gradient descent, without any further algorithmic or structural changes. Because of this, the model is fully task agnostic after deployment. SeqFOMAML may however be used in more complex setups in conjunction with other algorithms specific to continual learning. For example, the model may use Elastic Weight Consolidation \cite{kirkpatrick2017overcoming} as a regularizer to mitigate catastrophic forgetting, while at the same time optimizing the initial parameters using SeqFOMAML. The experiments presented here, however, use SeqFOMAML alone for simplicity and to show its contribution to mitigating catastrophic forgetting.

Let's take a sequence of tasks sampled from a distribution $\mathcal{T}_1, \dots, \mathcal{T}_n \sim p\left(\mathcal{T}_{1\dots n}\right)$. A model is sequentially trained on data (e.g., minibatches) sampled from a single task at a time, for a fixed number $k$ of iterations (per task). In this context, we wish for the model to maximize the objective of each task $i$ (e.g., minimize a loss function) both \textit{immediately after learning it} (i.e., $L_{\mathcal{T}_i, \text{test}}(\boldsymbol{\phi}^{\color{red}i})$ below), and \textit{after learning all tasks in the sequence} (i.e., $L_{\mathcal{T}_i, \text{test}}(\boldsymbol{\phi}^{\color{red}n})$): 


\begin{align}
    \minimize_{\boldsymbol{\phi}} \, \mathbb{E}_{\mathcal{T}_1, \dots, \mathcal{T}_n} \left[ \frac{1}{n} \sum_{i=1}^n {\color[rgb]{0.15,0.45,0.15} L_{i,\text{test}}\left( \boldsymbol{\phi}^n \right) }  +  {\color[rgb]{0.15,0.15,0.45} L_{i,\text{test}}\left( \boldsymbol{\phi}^i \right)} \right]
\label{eqn:SeqMAML_objective}
\end{align}

where $\boldsymbol{\phi}^j$ are the parameters obtained after sequentially training on all tasks until $j$, inclusive (e.g., $\boldsymbol{\phi}^n=U_{n,\text{train}}^k( U_{n-1,\text{train}}^k( \dots U_{1,\text{train}}^k(\boldsymbol{\phi}) \dots ) )$ ). Optimization is then performed by gradient descent:

\begin{align}
    &g_\text{SeqMAML} = \nabla_{\boldsymbol{\phi}} \mathbb{E}_{\mathcal{T}_1, \dots, \mathcal{T}_n} \left[ \frac{1}{n} \sum_{i=1}^n {\color[rgb]{0.15,0.45,0.15} L_{i,\text{test}}\left( \boldsymbol{\phi}^n \right) }  +  {\color[rgb]{0.15,0.15,0.45} L_{i,\text{test}}\left( \boldsymbol{\phi}^i \right)} \right] \nonumber \\
    &= \mathbb{E}_{\mathcal{T}_1, \dots, \mathcal{T}_n} \left[ \frac{1}{n} \sum_{i=1}^n {\color[rgb]{0.15,0.45,0.15} L_{i,\text{test}}'\left( \boldsymbol{\phi}^n \right) \frac{\partial \boldsymbol{\phi}^n}{\partial \boldsymbol{\phi}} }  +  {\color[rgb]{0.15,0.15,0.45} L_{i,\text{test}}'\left( \boldsymbol{\phi}^i \right) \frac{\partial \boldsymbol{\phi}^i}{\partial \boldsymbol{\phi}}} \right]
\label{eqn:SeqMAMLGradientLossAfterTask}
\end{align}

However, the number of inner gradient updates is $n$ times larger than the equivalent single-task setting in MAML (assuming a comparable number of per-task gradient updates), which makes the problem computationally expensive and the Jacobians harder to compute. It is thus necessary to use the first-order variant of the algorithm, FOMAML, that ignores second-order terms and replaces the Jacobian with the Identity matrix. The meta-gradient for the first-order variant becomes:

\begin{equation}
\label{eqn:SeqFOMAMLGradientLossAfterTask}
    g_\text{Seq\textbf{FO}MAML} = \mathbb{E}_{\mathcal{T}_1, \dots, \mathcal{T}_n} \left[ \frac{1}{n} \sum_{i=1}^n {\color[rgb]{0.15,0.45,0.15} L_{i,\text{test}}'\left( \boldsymbol{\phi}^n \right) }  +  {\color[rgb]{0.15,0.15,0.45} L_{i,\text{test}}'\left( \boldsymbol{\phi}^i \right) } \right] \\
\end{equation}


\section{Related Work}

The closest related work can be divided into two categories: variants of meta-continual learning, and evolution-based optimization of learning systems.

\textbf{Meta-continual learning.} One of the first approaches that exploited a meta-learning setting to address catastrophic forgetting is Risto et al. (2018) \cite{meta_cont_learning}, who proposed to learn an optimizer robust to catastrophic forgetting using a meta-dataset and pairs of adjacent tasks in a sequence. Closely related is the work by He et al. (2019) \cite{pascanu_taskagnostic_cl_via_metalearning}, who suggest to exploit meta-learning to speed up the \textit{recovery} of lost performance rather than focusing exclusively on remembering previous tasks, aided by the explicit inference of the current observed task.

Javed and White (2019) \cite{metalearning_features_for_cl} instead proposed a different approach, by meta-optimizing the network's parameters to learn useful features that can generalize well across tasks, thus minimizing interference as in our solution. Standard continual learning algorithms are then applied to the highest layers of the network, to explicitly address the problem of catastrophic forgetting in a traditional way. In a similar manner, Riemer et al. (2018) \cite{cl_max_transfer_min_interference} propose a method, Meta-Experience Replay (MER), based on meta-learning, that explicitly optimizes the network's parameters to maximize the alignment between tasks' gradients, thus reducing interference and improving positive transfer.


Finally, Finn et al. (2019) \cite{finn2019online} extend meta-learning to an online setting where tasks from the meta-distribution are presented sequentially. The work however does not focus on continual learning per se (catastrophic forgetting is mostly averted by using a memory buffer), but rather on forward transfer, and thus on making future learning increasingly faster and more efficient.

\textbf{Evolution-inspired meta-learning.} It has been suggested that meta-learning and evolution share a crucial connection, via the Baldwin effect \cite{baldwinian_meta_learning}. However, they have also been linked in more implicit ways by using evolutionary algorithms to optimize plastic neural networks (for a complete review, see \cite{soltoggio2018born}).

Close to the scope of this paper is the work of Ellefsen et al. (2015) \cite{ellefsen2015neural}, that proposes to optimize the connections of a neural network by evolution, with an added cost per connection. Continual learning is possible due to the emergence of modularity in the network and segregation between the tasks. Crucially, agents are trained with a concept of a ``lifetime'', during which they are exposed to a non-stationary environment that forces them to remember previously acquired skills and knowledge. While the learning rule was not learnt, separate works have explored such possibility, for example in Backpropamine \cite{backpropamine}, where gradient descent is performed on a differentiable Hebbian learning rule in a meta-learning context.

\section{Experiments}
\label{sec:experiments}

In this section we evaluate SeqFOMAML on two benchmarks, split-Omniglot and split-MiniImageNet.

All the experiments were implemented in Python using Tensorflow 2.0 \cite{tensorflow}, and most of the code is publicly available as part of the pyMeta library \footnote{\url{https://github.com/spiglerg/pyMeta}}. The base network architectures and parameters of the experiments were adapted from related work \cite{maml, reptile}, and are reported in the Supplementary Materials. The experiments were run on a single computer (nVidia RTX 2080 Super).

\subsection{Split-Omniglot (single-headed network)}

In the first experiment, SeqFOMAML is evaluated on an incremental task learning scenario \cite{hsu2018re} using the split-Omniglot benchmark, in which a neural network is trained on a sequence of classification tasks built from different subsets of Omniglot characters \cite{omniglot}, without the possibility of storing or accessing previously observed data from the sequence.

Testing was done as follows: first, the network was pre-meta-trained on the meta-train dataset using SeqFOMAML with sequences of different length for each experiment (i.e., $n$ in Eqs. \ref{eqn:SeqMAML_objective} and \ref{eqn:SeqFOMAMLGradientLossAfterTask}; $L=1$, equivalent to vanilla FOMAML, $L=5$, $L=10$ and $L=20$). Next, the network was tested on the split-Omniglot benchmark by sequential training on test sequences composed by $100$ $5$-way tasks sampled from the meta-test set, and its accuracy on all previously observed tasks was recorded after training on each new task in the sequence. Note that during evaluation (after pre-meta-training using SeqFOMAML) the network is fully task-agnostic.

\begin{figure*}[!t]
\centering
\includegraphics[width=\linewidth]{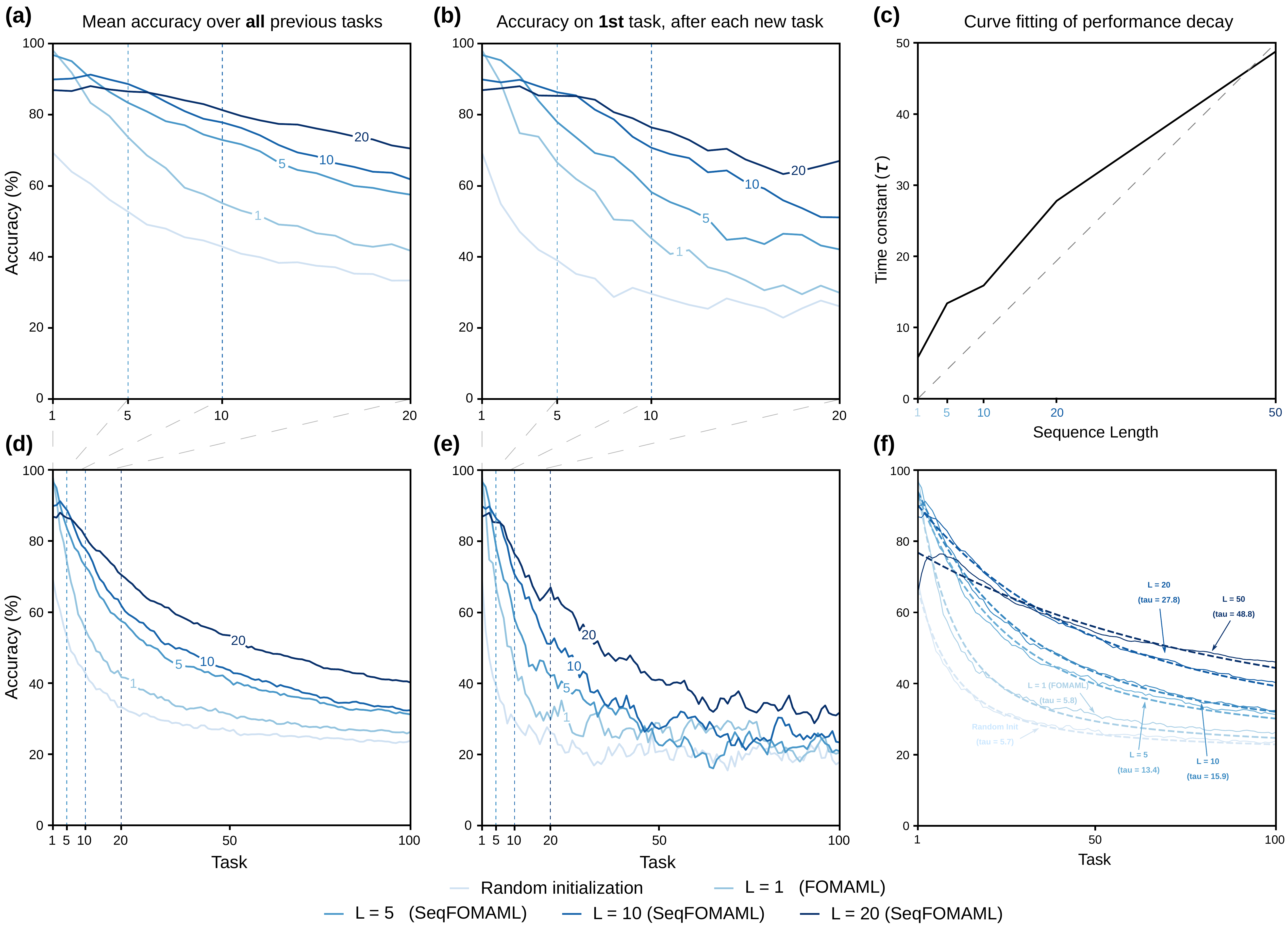}


\caption{ Split-Omniglot: sequential learning of $100$ $5$-way Omniglot tasks \textbf{(d, e)}, and close-up of the first $20$ in the sequence \textbf{(a, b)}, averaged over $100$ runs. After training on each task, the mean accuracy on all previous tasks is computed \textbf{(a, d)}. \textbf{(b, e)} show the accuracy of the network on the first task, repeatedly tested after sequential training on each task in the sequence. Vertical dashed lines indicate the length of the sequences used during meta-training for the different models (lengths $L=\{5, 10, 20\}$ tasks, along with $L=1$ equivalent to vanilla FOMAML). \textbf{(f)}: the performance on each task throughout the sequence is modeled as an exponential decay, in order to estimate the rate of decay (forgetting) for the different models. \textbf{(c)}: the fitted time constants are plotted as a function of the corresponding SeqFOMAML sequence length. }

\label{fig:results_by_length}
\end{figure*}

Figure \ref{fig:results_by_length} shows a comparison of SeqFOMAML pre-meta-trained using sequences of different lengths, and a baseline randomly initialized network (i.e., non meta-trained). In particular, models trained with a fixed sequence length $L$ should be optimized for a similar test-time sequence length, outside of which good performance is due to meta-generalization to longer sequences that were not observed during training.

Finally, the decay in performance on each trained task was modeled as an exponential decay function in order to estimate the rate of forgetting. The rate of decay of the per-task performance is estimated by fitting a function of the predicted mean accuracy tested on all previous tasks, after each task $t$ (i.e., the curves in Figure \ref{fig:results_by_length}, left column):

$$
F(t) = \frac{1}{t} \sum_{i \leq t} f_i(t)
$$

where the accuracy $f_i(t)$ of each task $i$, tested immediately after training on task $t$ ($t \geq i$) is assumed to decay exponentially during successive training, with an asymptotic chance-level performance of $20\%$ for $5$-way Omniglot classification tasks
$$
f_i(t) = a e^{-\frac{1}{\tau} (t-i)} + 0.2
$$

The parameters $a$ and $\tau$ were fitted using the Trust Region Reflective algorithm from SciPy (\texttt{scipy.optimize.curve\_fit}).

The fitted curves $F(t)$ are shown in Figure \ref{fig:results_by_length}.\textbf{f}, along with a plot (\ref{fig:results_by_length}.\textbf{c}) of the fitted time constants $\tau$ as a function of the SeqFOMAML sequence length $L$ ($\tau=\{5.8, 13.4, 15.9, 27.8, 48.8\}$ respectively for $L=\{1\text{(baseline FOMAML)}, 5, 10, 20, 50\}$). The time constants were found to be strongly correlated to the length of the training sequences ($r=0.99, n=5, p<0.01$).

\subsection{Split-MiniImageNet (multi-headed network)}

The second experiment had the objective of both evaluating SeqFOMAML on a more challenging scenario, and of exploring the effect of ``good'', reusable features learnt by standard (non-sequential) meta-learning \cite{bengio_maml,matthew2020}, in mitigating catastrophic forgetting \cite{mcrae1993catastrophic,metalearning_features_for_cl,french1999catastrophic}.

SeqFOMAML is evaluated on a split-MiniImageNet benchmark, constructed in a manner similar to split-Omniglot, but built on the more complex MiniImageNet dataset \cite{miniimagenet1,miniimagenet2}. In this evaluation, SeqFOMAML is tested for incremental \emph{class} learning \cite{hsu2018re}, and as such it makes use of a multi-headed network, each head computing a separate softmax operation. The number of heads is not fixed, but rather it is dynamically incremented by adding new output layers initialized with the initial meta-trained parameters (SeqFOMAML) of the first head. As such, and contrary to the first experiment, the network is \emph{not} task-agnostic.

\begin{figure*}[!ht]
\centering
\includegraphics[width=0.7\linewidth]{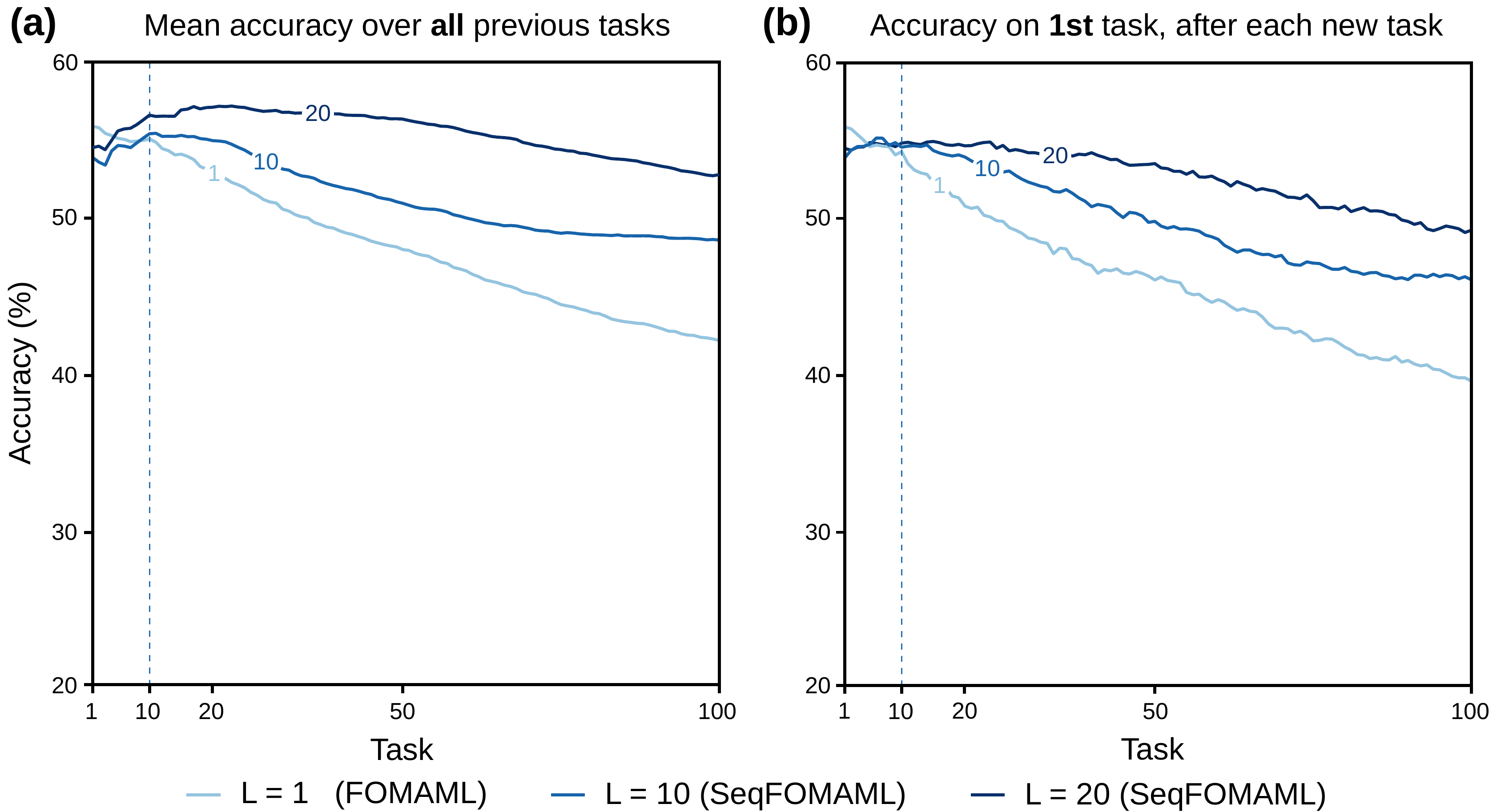}

\caption{ Split-MiniImageNet: sequential learning of $100$ $5$-way MiniImageNet tasks, averaged over $30$ runs. \textbf{(a)}: after training on each task, the mean accuracy on all previous tasks is computed. \textbf{(b)}: the accuracy of the network on the first task, tested after sequential training on each task in the sequence. The vertical dashed lines indicate the length of the sequences used during meta-training. The performance of each task during training is modeled as an exponential decay which is fitted to the data from the simulations to estimate the rate of decay (fitted time constants $\tau_1=91.5$ for $L=1$, $\tau_{10}=200.9$ for $L=10$, and $\tau_{10}=273.8$ for $L=20$). }

\label{fig:mh_results_by_length}
\end{figure*}

Figure \ref{fig:mh_results_by_length} shows a comparison of the results for SeqFOMAML ($L=10$ and $L=20$) and a baseline FOMAML model ($L=1$). The decay in performance was fitted as in split-Omniglot, estimating a time constant of $\tau_1=91.5$ for the baseline model and $\tau_{10}=200.9, \tau_{20}=273.8$ for the SeqFOMAML model. While SeqFOMAML achieves a significantly lower rate of decay,  the high performance of the baseline suggests that in a multi-headed setup, even the representations learnt by standard meta-learning (FOMAML) are sufficient to mitigate catastrophic forgetting (compare, e.g., to the fast rate of forgetting of the baseline in Figure \ref{fig:results_by_length} ).

\section{Discussion and Conclusion}

We show that catastrophic forgetting can be mitigated in a meta-learning context, by exposing a model to multiple tasks during training, rather than a single task as in traditional paradigms. Specifically, and contrary to vanilla meta-learning approaches, full sequences of tasks are sampled from a distribution, instead of sampling a single task at a time. We further present SeqFOMAML, a meta-learning algorithm designed to jointly optimize for fast adaptation to new tasks, as in traditional single-task meta-learning, and for performance on previously observed tasks. Notably, SeqFOMAML learns a prior for sequential task learning in the form of initial model parameters, as in traditional meta-learning, but after that the actual continual learning over a sequence of tasks \emph{relies on plain stochastic gradient descent}, without any further change to the model or to the training regime.

The approach presented shares most of the properties of traditional meta-learning. In particular, it work by fine-tuning pre-meta-trained parameters that have been optimized for generalization over tasks from a given distribution. Contrary to non-sequential meta-learning, however, SeqFOMAML explicitly optimizes for \emph{remembering} information about tasks that are observed in a sequence, rather than only optimizing for fast learning on new tasks.

SeqFOMAML was shown to significantly reduce the amount of catastrophic interference between tasks on two benchmarks, split-Omniglot and split-MiniImageNet, which was further confirmed by estimating the time constants of the decay of the accuracy of the models on tasks in the observed sequence.

Nonetheless, it was found that the performance of a multi-headed network initialized with parameters obtained from standard meta-training (FOMAML) exhibited a low rate of decay, even without specific provisions to mitigate catastrophic forgetting (see Figure \ref{fig:mh_results_by_length}). This may be due to the type of representations learnt by (FO)MAML, that have been shown to be optimized for re-use across the different tasks in the meta-training distribution \cite{bengio_maml,matthew2020}, and have been already exploited explicitly in a continual learning context \cite{metalearning_features_for_cl}.

The result would then be in line with foundational work in the field by McRae and French, who respectively observed that catastrophic forgetting could be mitigated by pre-training a network to help it develop hidden representations useful to all the tasks in a sequence, rather than to a single current task \cite{mcrae1993catastrophic}, and that different types of learnt features affect subsequent catastrophic interference by different degrees \cite{french1999catastrophic}.

While SeqFOMAML was found to perform well in the experiments we presented, a concern may arise on its capacity to scale to larger and more complex problems. This problem is shared in general with most meta-learning based approaches to continual learning, as well as evolutionary approaches that require the use of long sequences of tasks or experiences. Notably, however, even short tasks sequences were found to be helpful in mitigating catastrophic forgetting, so training a model on full long sequences may not be always required, or a form of curriculum learning may be used to speed-up the initial meta-training. It is also important to note that the computational burden is restricted to the initial pre-meta-training, after which the model can be deployed and fine-tuned on the actual desired sequence of tasks by plain stochastic gradient descent.

Finally, it is useful to note that SeqFOMAML is model-agnostic, so it can in principle be used with any other continual learning method.

Future work should focus on interfacing SeqFOMAML with effective complementary continual learning methods like Elastic Weight Consolidation \cite{kirkpatrick2017overcoming}, to investigate the effectiveness of such synergies.

\bibliography{references}

\begin{thebibliography}{30}
\providecommand{\natexlab}[1]{#1}
\providecommand{\url}[1]{\texttt{#1}}
\expandafter\ifx\csname urlstyle\endcsname\relax
  \providecommand{\doi}[1]{doi: #1}\else
  \providecommand{\doi}{doi: \begingroup \urlstyle{rm}\Url}\fi

\bibitem[Abadi et~al.(2016)Abadi, Agarwal, Barham, Brevdo, Chen, Citro,
  Corrado, Davis, Dean, Devin, et~al.]{tensorflow}
Abadi, M., Agarwal, A., Barham, P., Brevdo, E., Chen, Z., Citro, C., Corrado,
  G.~S., Davis, A., Dean, J., Devin, M., et~al.
\newblock Tensorflow: Large-scale machine learning on heterogeneous distributed
  systems.
\newblock \emph{arXiv preprint arXiv:1603.04467}, 2016.

\bibitem[Andrychowicz et~al.(2016)Andrychowicz, Denil, Gomez, Hoffman, Pfau,
  Schaul, Shillingford, and De~Freitas]{ml_graddesc_bygraddesc_2016}
Andrychowicz, M., Denil, M., Gomez, S., Hoffman, M.~W., Pfau, D., Schaul, T.,
  Shillingford, B., and De~Freitas, N.
\newblock Learning to learn by gradient descent by gradient descent.
\newblock In \emph{Advances in neural information processing systems}, pp.\
  3981--3989, 2016.

\bibitem[Baldwin(1896)]{baldwin_effect_original}
Baldwin, J.~M.
\newblock A new factor in evolution.
\newblock \emph{The american naturalist}, 30\penalty0 (354):\penalty0 441--451,
  1896.

\bibitem[Bengio et~al.(1992)Bengio, Bengio, Cloutier, and
  Gecsei]{metalearning_bengio1992optimization}
Bengio, S., Bengio, Y., Cloutier, J., and Gecsei, J.
\newblock On the optimization of a synaptic learning rule.
\newblock In \emph{Preprints Conf. Optimality in Artificial and Biological
  Neural Networks}, pp.\  6--8. Univ. of Texas, 1992.

\bibitem[Ellefsen et~al.(2015)Ellefsen, Mouret, and Clune]{ellefsen2015neural}
Ellefsen, K.~O., Mouret, J.-B., and Clune, J.
\newblock Neural modularity helps organisms evolve to learn new skills without
  forgetting old skills.
\newblock \emph{PLoS computational biology}, 11\penalty0 (4):\penalty0
  e1004128, 2015.

\bibitem[Fernando et~al.(2018)Fernando, Sygnowski, Osindero, Wang, Schaul,
  Teplyashin, Sprechmann, Pritzel, and Rusu]{baldwinian_meta_learning}
Fernando, C., Sygnowski, J., Osindero, S., Wang, J., Schaul, T., Teplyashin,
  D., Sprechmann, P., Pritzel, A., and Rusu, A.
\newblock Meta-learning by the baldwin effect.
\newblock In \emph{Proceedings of the Genetic and Evolutionary Computation
  Conference Companion}, pp.\  1313--1320. ACM, 2018.

\bibitem[Finn et~al.(2017)Finn, Abbeel, and Levine]{maml}
Finn, C., Abbeel, P., and Levine, S.
\newblock Model-agnostic meta-learning for fast adaptation of deep networks.
\newblock In \emph{Proceedings of the 34th International Conference on Machine
  Learning-Volume 70}, pp.\  1126--1135. JMLR. org, 2017.

\bibitem[Finn et~al.(2019)Finn, Rajeswaran, Kakade, and Levine]{finn2019online}
Finn, C., Rajeswaran, A., Kakade, S., and Levine, S.
\newblock Online meta-learning.
\newblock \emph{arXiv preprint arXiv:1902.08438}, 2019.

\bibitem[French(1999)]{french1999catastrophic}
French, R.~M.
\newblock Catastrophic forgetting in connectionist networks.
\newblock \emph{Trends in cognitive sciences}, 3\penalty0 (4):\penalty0
  128--135, 1999.

\bibitem[Goodfellow et~al.(2013)Goodfellow, Mirza, Xiao, Courville, and
  Bengio]{goodfellow2013empirical}
Goodfellow, I.~J., Mirza, M., Xiao, D., Courville, A., and Bengio, Y.
\newblock An empirical investigation of catastrophic forgetting in
  gradient-based neural networks.
\newblock \emph{arXiv preprint arXiv:1312.6211}, 2013.

\bibitem[He et~al.(2015)He, Zhang, Ren, and Sun]{human_level_imagenet}
He, K., Zhang, X., Ren, S., and Sun, J.
\newblock Delving deep into rectifiers: Surpassing human-level performance on
  imagenet classification.
\newblock In \emph{Proceedings of the IEEE international conference on computer
  vision}, pp.\  1026--1034, 2015.

\bibitem[He et~al.(2019)He, Sygnowski, Galashov, Rusu, Teh, and
  Pascanu]{pascanu_taskagnostic_cl_via_metalearning}
He, X., Sygnowski, J., Galashov, A., Rusu, A.~A., Teh, Y.~W., and Pascanu, R.
\newblock Task agnostic continual learning via meta learning.
\newblock \emph{arXiv preprint arXiv:1906.05201}, 2019.

\bibitem[Hinton \& Nowlan(1987)Hinton and
  Nowlan]{baldwin_effect_neuralnetworks}
Hinton, G.~E. and Nowlan, S.~J.
\newblock How learning can guide evolution.
\newblock \emph{Complex systems}, 1\penalty0 (3):\penalty0 495--502, 1987.

\bibitem[Hsu et~al.(2018)Hsu, Liu, Ramasamy, and Kira]{hsu2018re}
Hsu, Y.-C., Liu, Y.-C., Ramasamy, A., and Kira, Z.
\newblock Re-evaluating continual learning scenarios: A categorization and case
  for strong baselines.
\newblock \emph{arXiv preprint arXiv:1810.12488}, 2018.

\bibitem[Javed \& White(2019)Javed and White]{metalearning_features_for_cl}
Javed, K. and White, M.
\newblock Meta-learning representations for continual learning.
\newblock \emph{arXiv preprint arXiv:1905.12588}, 2019.

\bibitem[Kirkpatrick et~al.(2017)Kirkpatrick, Pascanu, Rabinowitz, Veness,
  Desjardins, Rusu, Milan, Quan, Ramalho, Grabska-Barwinska,
  et~al.]{kirkpatrick2017overcoming}
Kirkpatrick, J., Pascanu, R., Rabinowitz, N., Veness, J., Desjardins, G., Rusu,
  A.~A., Milan, K., Quan, J., Ramalho, T., Grabska-Barwinska, A., et~al.
\newblock Overcoming catastrophic forgetting in neural networks.
\newblock \emph{Proceedings of the national academy of sciences}, 114\penalty0
  (13):\penalty0 3521--3526, 2017.

\bibitem[Lake et~al.(2011)Lake, Salakhutdinov, Gross, and Tenenbaum]{omniglot}
Lake, B., Salakhutdinov, R., Gross, J., and Tenenbaum, J.
\newblock One shot learning of simple visual concepts.
\newblock In \emph{Proceedings of the annual meeting of the cognitive science
  society}, volume~33, 2011.

\bibitem[McCloskey \& Cohen(1989)McCloskey and
  Cohen]{mccloskey1989catastrophic}
McCloskey, M. and Cohen, N.~J.
\newblock Catastrophic interference in connectionist networks: The sequential
  learning problem.
\newblock In \emph{Psychology of learning and motivation}, volume~24, pp.\
  109--165. Elsevier, 1989.

\bibitem[McRae \& Hetherington(1993)McRae and
  Hetherington]{mcrae1993catastrophic}
McRae, K. and Hetherington, P.~A.
\newblock Catastrophic interference is eliminated in pretrained networks.
\newblock In \emph{Proceedings of the 15h annual conference of the cognitive
  science society}, pp.\  723--728, 1993.

\bibitem[Miconi et~al.(2018)Miconi, Rawal, Clune, and Stanley]{backpropamine}
Miconi, T., Rawal, A., Clune, J., and Stanley, K.~O.
\newblock Backpropamine: training self-modifying neural networks with
  differentiable neuromodulated plasticity.
\newblock 2018.

\bibitem[Nichol et~al.(2018)Nichol, Achiam, and Schulman]{reptile}
Nichol, A., Achiam, J., and Schulman, J.
\newblock On first-order meta-learning algorithms.
\newblock \emph{arXiv preprint arXiv:1803.02999}, 2018.

\bibitem[Raghu et~al.(2019)Raghu, Raghu, Bengio, and Vinyals]{bengio_maml}
Raghu, A., Raghu, M., Bengio, S., and Vinyals, O.
\newblock Rapid learning or feature reuse? towards understanding the
  effectiveness of maml.
\newblock \emph{arXiv preprint arXiv:1909.09157}, 2019.

\bibitem[Ravi \& Larochelle(2016)Ravi and Larochelle]{miniimagenet2}
Ravi, S. and Larochelle, H.
\newblock Optimization as a model for few-shot learning.
\newblock 2016.

\bibitem[Riemer et~al.(2018)Riemer, Cases, Ajemian, Liu, Rish, Tu, and
  Tesauro]{cl_max_transfer_min_interference}
Riemer, M., Cases, I., Ajemian, R., Liu, M., Rish, I., Tu, Y., and Tesauro, G.
\newblock Learning to learn without forgetting by maximizing transfer and
  minimizing interference.
\newblock \emph{arXiv preprint arXiv:1810.11910}, 2018.

\bibitem[Schmidhuber(1987)]{metalearning_schmidhuber1987}
Schmidhuber, J.
\newblock \emph{Evolutionary principles in self-referential learning, or on
  learning how to learn: the meta-meta-... hook}.
\newblock PhD thesis, Technische Universit{\"a}t M{\"u}nchen, 1987.

\bibitem[Silver et~al.(2016)Silver, Huang, Maddison, Guez, Sifre, Van
  Den~Driessche, Schrittwieser, Antonoglou, Panneershelvam, Lanctot,
  et~al.]{alphago}
Silver, D., Huang, A., Maddison, C.~J., Guez, A., Sifre, L., Van Den~Driessche,
  G., Schrittwieser, J., Antonoglou, I., Panneershelvam, V., Lanctot, M.,
  et~al.
\newblock Mastering the game of go with deep neural networks and tree search.
\newblock \emph{nature}, 529\penalty0 (7587):\penalty0 484, 2016.

\bibitem[Soltoggio et~al.(2018)Soltoggio, Stanley, and Risi]{soltoggio2018born}
Soltoggio, A., Stanley, K.~O., and Risi, S.
\newblock Born to learn: the inspiration, progress, and future of evolved
  plastic artificial neural networks.
\newblock \emph{Neural Networks}, 108:\penalty0 48--67, 2018.

\bibitem[Srivatanapa(2020)]{matthew2020}
Srivatanapa, V.
\newblock A quantitative and qualitative analysis of first-order model-agnostic
  meta-learning.
\newblock Master's thesis, Tilburg University, 2020.

\bibitem[Vinyals et~al.(2016)Vinyals, Blundell, Lillicrap, Wierstra,
  et~al.]{miniimagenet1}
Vinyals, O., Blundell, C., Lillicrap, T., Wierstra, D., et~al.
\newblock Matching networks for one shot learning.
\newblock In \emph{Advances in neural information processing systems}, pp.\
  3630--3638, 2016.

\bibitem[Vuorio et~al.(2018)Vuorio, Cho, Kim, and Kim]{meta_cont_learning}
Vuorio, R., Cho, D.-Y., Kim, D., and Kim, J.
\newblock Meta continual learning.
\newblock \emph{arXiv preprint arXiv:1806.06928}, 2018.

\end{thebibliography}
\bibliographystyle{icml2020}

\onecolumn

\appendix

\section{Methods}

\subsection{pyMeta and Open Source implementation}

Meta-training over sequences of tasks instead of individual tasks adds a further layer of complexity to the problem. For this reason, we have made a \textbf{pyMeta} library, based on Tensorflow \cite{tensorflow}, publicly available (\url{https://github.com/spiglerg/pyMeta}), to simplify the definition of distributions over tasks (classfication, regression and RL) and sequences of tasks, and to allow sampling from them.

\textbf{pyMeta} is organized on the concept of Tasks, for example ``ClassificationTask'', that can be used to implement a variety of machine learning problems. Tasks can then be grouped within ``TaskDistribution'' objects that allow to define distributions over tasks that can be sampled from. The popular meta-learning algorithms FOMAML and Reptile are also included as baselines.

Most of the code used in this paper is made available in the pyMeta repository.

\subsection{Network architecture and parameters of the simulations}

The parameters of the simulations were adapted from related work \cite{maml, reptile} and partially tuned using the validation set. The network architecture is a standard CNN used for few-shot Omniglot classification \cite{maml}, and is composed by $4$ blocks, each comprising a convolutional layer with $64$ filters of size $3 \times 3$, stride $1$, followed by batch normalization, $2 \times 2$ max pooling and ReLU activation. The rest of the hyperparameters are reported in Table \ref{tbl:hyperparameters_5way_omniglot}.

\begin{table}[h!]
\renewcommand{\arraystretch}{1.3}
\caption{Hyperparameters for the simulations on the Omniglot dataset.}
\label{tbl:hyperparameters_5way_omniglot}
\centering
  \begin{tabular}{ |c|c| }
    \hline
    Parameter & 5-way \\ \hline \hline
    Inner batch size & 5*10 \\
    Inner optimizer & \textit{Plain SGD} \\
    Inner learning rate & 0.01 \\
    \# train samples per class & 10 \\
    \# test samples per class & 10 \\
    \# inner iterations & 5 \\
    \hline
    Meta batch size & 5 \\
    Meta optimizer & \textit{Adam} \\
    Meta learning rate & 0.001 \\
    \# outer meta-training iterations & 20000 \\
    \hline
  \end{tabular}
\renewcommand{\arraystretch}{1}
\end{table}

\section{Supplementary Results}

\subsection{Fast adaptation versus positive backward transfer}

We further explore the balance in SeqFOMAML between \textit{fast learning and adaptation} to new tasks in the sequence -as in traditional meta-learning, and possibly with the aid of positive forward transfer between tasks,- and \textit{positive backward transfer}.

To do so, we train the same sequential learning problem (sequences of $5$-way Omniglot classification tasks, $L=10$ tasks per training sequence) using two different objectives. In one case, SeqFOMAML is used as described in Section 2, and specifically optimization is performed with the SeqFOMAML objective, which aims at minimizing the loss function of each task in the sequence both at the end of training, at parameters $\boldsymbol{\phi}^n$, and immediately after training on each task $i$, $\boldsymbol{\phi}^i$. The second case instead exclusively optimizes performance of each task at the end of training, i.e. at parameters $\boldsymbol{\phi}^n$.

\begin{figure}[!t]
\centering
\includegraphics[width=0.9\linewidth]{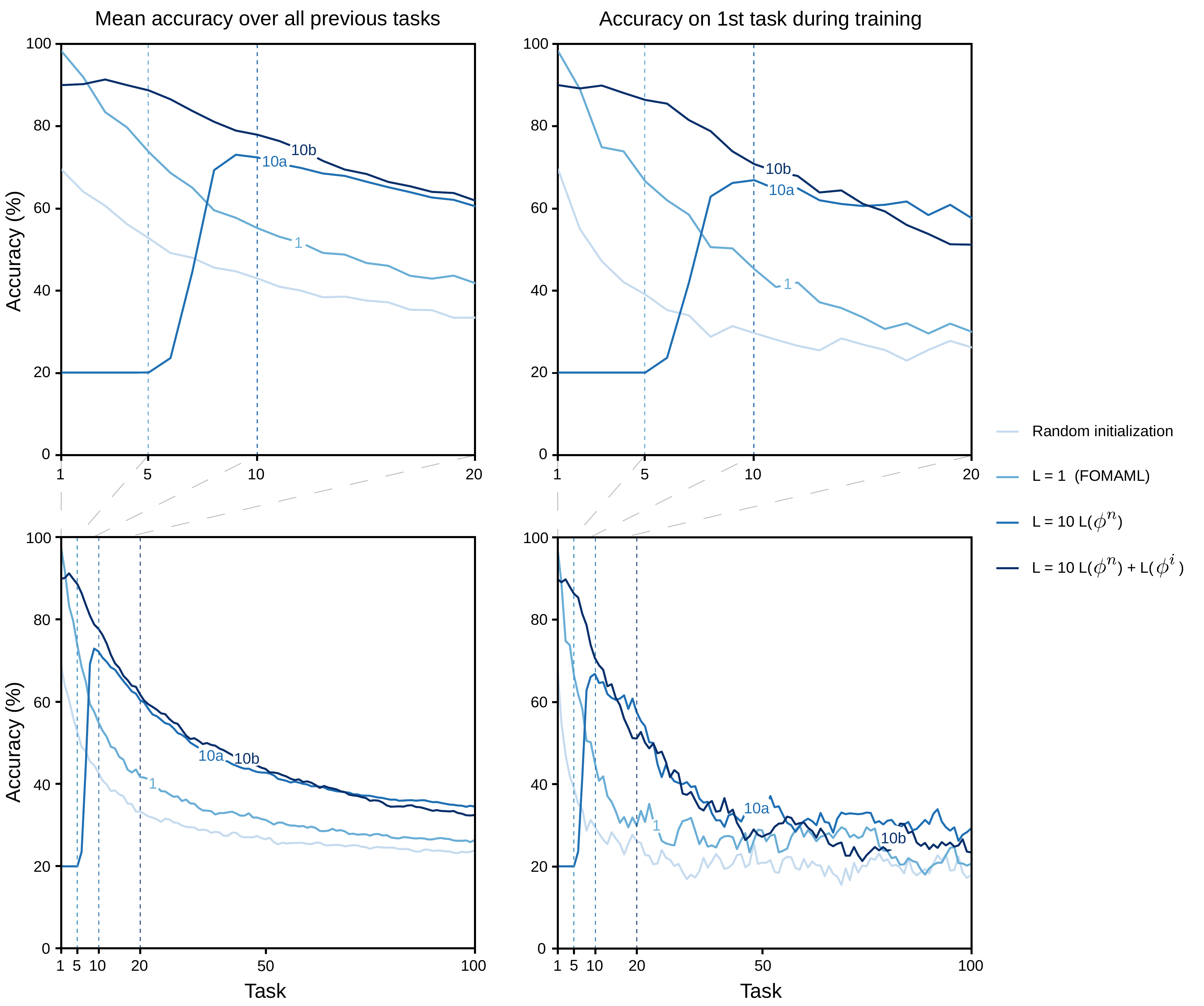}

\caption{ The same simulations as in the main text are replicated with a fixed length of $L=10$ training tasks, in two variants of SeqFOMAML. One case ($\mathbf{10b}$) is equivalent to the simulations in the main text, where the objective optimized by SeqFOMAML is the same as the SeqMAML objective, that minimizes the loss of each task in the sequence both at the final parameters $\boldsymbol{\phi}^n$ and after training on each task $i$, $\boldsymbol{\phi}^i$. The other case ($\mathbf{10a}$) instead optimizes the performance on each task exclusively at the end of the sequence $\boldsymbol{\phi}^n$, thus not explicitly optimizing for fast learning and adaptation to new tasks, but only relying on positive backward transfer. }

\label{fig:results_by_loss}
\end{figure}

The results are shown in Figure \ref{fig:results_by_loss}. Because performance is not optimized to be high immediately after training on each task, the model performs poorly when exposed to few tasks, as performance is driven only by positive backward transfer, so that training on new tasks improves the performance of the model on previous ones.


\end{document}